\documentclass[runningheads]{llncs}
\usepackage{graphicx}
\usepackage{amsmath,amssymb} % define this before the line numbering.
\usepackage{color}
\newcommand*\rot{\rotatebox{90}}
\usepackage{array}
\usepackage{booktabs}
\usepackage{pifont}
\usepackage[width=122mm,left=12mm,paperwidth=146mm,height=193mm,top=12mm,paperheight=217mm]{geometry}

\usepackage[table,xcdraw]{xcolor}
\usepackage{booktabs}
\usepackage{amssymb}
\usepackage{subfig}
\usepackage{xspace}
\usepackage{comment}
\newcommand*{\eg}{\emph{e.g.}\@\xspace}
\newcommand*{\ie}{\emph{i.e.}\@\xspace}
\newcommand*{\etal}{\emph{et al.}\@\xspace}
\newcommand*{\etc}{\emph{etc.}\@\xspace}

\begin{document}
% \renewcommand\thelinenumber{\color[rgb]{0.2,0.5,0.8}\normalfont\sffamily\scriptsize\arabic{linenumber}\color[rgb]{0,0,0}}
% \renewcommand\makeLineNumber {\hss\thelinenumber\ \hspace{6mm} \rlap{\hskip\textwidth\ \hspace{6.5mm}\thelinenumber}}
% \linenumbers
\pagestyle{headings}
\mainmatter

\title{Hard Negative Mining for \\Metric Learning Based Zero-Shot Classification} % Replace with your title

\titlerunning{Hard Negative Mining for Metric Learning Based Zero-Shot Classification}

\authorrunning{Bucher, Herbin and Jurie}

\author{Maxime Bucher$^{1,2}$, St\'ephane Herbin$^1$, Fr\'ed\'eric Jurie$^2$}

%Please write out author names in full in the paper, i.e. full given and family names. 
%If any authors have names that can be parsed into FirstName LastName in multiple ways, please include the correct parsing, in a comment to the volume editors:
%\index{Lastnames, Firstnames}
%(Do not uncomment it, because you may introduce extra index items if you do that...)

\institute{$^1$ONERA - The French Aerospace Lab, Palaiseau, France \\
$^2$Normandie Univ, UNICAEN, ENSICAEN, CNRS, Caen, France}

\maketitle
\begin{abstract}
Zero-Shot learning has been shown to be an efficient strategy for domain adaptation. In this context, this paper builds on the recent work of Bucher \etal \cite{bucher:hal-01348827}, which proposed an approach to solve Zero-Shot classification problems (ZSC) by introducing a novel metric learning based objective function. This objective function allows to learn an optimal embedding of the attributes jointly with a measure of similarity between images and attributes. This paper extends their approach by proposing several schemes to control the generation of the negative pairs, resulting in a significant improvement of the performance and giving above state-of-the-art results on three challenging ZSC datasets.  

\end{abstract}

\section{Introduction}

Among the different image interpretation methods exploiting some kind of knowledge transfer in their design, Zero Shot Classification (ZSC) can be considered as a domain adaptation problem where the new target domain is defined using an intermediate level of representation made of human understandable {\em semantic attributes}. The source domain is defined by an annotated image database expected to capture the relation between data and attribute based representation of classes.

Most of the recent approaches addressing ZSC \cite{Lampert:2014fs,Akata:2015tv,romera2015embarrassingly,Zhang:2015vs,zhang2016zero,wang2016zero,xian2016latent} rely on the computation of a similarity function in the semantic space. They learn the semantic embedding, either from data or from class description, and compare the embedded data using standard distance. 

Recently, \cite{bucher:hal-01348827} proposed to add a metric learning (ML) step to adapt empirically the similarity distance in the embedding space, leading to a multi-objective criterion optimizing both the metric and the embedding. The metric is learned using an empirical optimized criterion on random but equally sampled pairs of similar (positive) and dissimilar (negative) data.

In this paper, following observations from the active learning community (see \eg, \cite{Fu2013cu}),  we show that a careful choice of the negative pairs combined with the multi-objective criterion proposed in \cite{bucher:hal-01348827} leads to above state of the art results.

\section{Improved ZSL by efficient hard negative mining\label{sec:method}}

In a recent work, Bucher \etal \cite{bucher:hal-01348827} introduced a metric learning step in their zero-shot classification pipeline. Their model is trained from pairs of data, where positive (resp. negative) pairs are obtained by taking  the training images associated with their own provided attribute vector (resp. by  randomly assigning  attribute vector of another image) and are assigned to the class label '1' (resp. `-1'). The set of positive pairs are denoted $\mathbf{D_{+}}$ in the following, and $\mathbf{D_{-}}$ for the negatives. This paper investigates improved ways to select negative pairs. 

\subsection{Bucher \etal metric learning framework for zero-shot classification}
This section summarizes  Bucher \etal~\cite{bucher:hal-01348827} paper, to which the reader should refer for more details. Zero-shot classification problem is cast into an optimal framework of the form:
\[ \mathbf{Y}^* = \operatorname*{arg\,min}_{\mathbf{Y}\in\mathcal{Y}} S(\mathbf{X},\mathbf{Y}), \]
where $\mathbf{X}$ is an image, $\mathbf{Y}$ a vector of attributes and $\mathbf{S}$ a parametric similarity measure. A metric matrix denoted as $\mathbf{W}_A$ transforms the attribute  embedding space into a space where the Euclidean distance can be used. The similarity between images and attributes is computed as:

\begin{equation}
S(\mathbf{X},\mathbf{Y}) = \left\Vert(\hat{\mathbf{A}}_X(\mathbf{X}) - \mathbf{Y})^T \mathbf{W}_A \right\Vert_2
\label{eq:score_attribute_embedding}
\end{equation}

\noindent where $\hat{\mathbf{A}}_X$ embeds the $\mathbf{X}$ modality into the space of $\mathbf{Y}$ using a linear transformation combined with a ReLU type transfer function:
\begin{equation}
\hat{\mathbf{A}}_X(\mathbf{X}) = \max(0,\mathbf{X}^T\mathbf{W}_X  + \mathbf{b}_X).
\label{eq:embedding_attribute}
\end{equation}

The role of learning is to estimate jointly the two matrices $\mathbf{W}_X$ and $\mathbf{W}_A$. The empirical learning criterion used is the sum of 3 terms: 

(i) a term for the metric $\mathbf{W}_A$:
\begin{equation}
l_H(\mathbf{X}_i,\mathbf{Y}_i,Z_i,\tau) = \max\left(0,1-Z_i(\tau - S(\mathbf{X}_i,\mathbf{Y}_i)^2)\right).
\label{eq:criterion_hinge_loss}
\end{equation}
where $Z_i$ states that the two modalities are consistent ($Z_i=1$) or not ($Z_i=-1$). $\tau$ is the threshold separating similar and dissimilar examples.

(ii) a quadratic loss for the linear attribute prediction $\mathbf{W}_X$ (only applied to positive pairs):

\begin{equation}
l_A(\mathbf{X}_i,\mathbf{Y}_i,Z_i) = \max(0,Z_i).\left\Vert \mathbf{Y}_i - \hat{\mathbf{A}}_X(\mathbf{X}_i) \right\Vert_2^2.
\label{eq:criterion_attribute_prediction}
\end{equation}

(iii) a quadratic penalization to prevent overfitting: 
\begin{equation}
R(\mathbf{W}_A,\mathbf{W}_X, \mathbf{b}_X) = \left\Vert \mathbf{W}_X \right\Vert_F^2 + \left\Vert \mathbf{b}_X \right\Vert_2^2 + \left\Vert \mathbf{W}_A \right\Vert_F^2 
\label{eq:criterion_weight_regularization}
\end{equation}

The overall objective function can then now be written as the sum of the previously defined terms:
\begin{equation}
\begin{split}
\mathcal{L}(\mathbf{W}_A,\mathbf{W}_X, \mathbf{b}_X, \tau) = \sum_i l_H(\mathbf{X_i},\mathbf{Y_i},Z_i,\tau) + \lambda \sum_i l_A(\mathbf{X_i},\mathbf{Y_i},Z_i)\\
 + \mu R(\mathbf{W}_A,\mathbf{W}_X, \mathbf{b}_X)
\label{eq:criterion_global_bucher}
\end{split}
\end{equation}

In this context, ZSC is achieved by finding the most consistent attribute description from a set of exclusive attribute class descriptors $\{\mathbf{Y}^*_k\}_{k=1}^C$ given the image where $k$, is the index of a class:

\begin{equation}
k^* = \operatorname*{arg\,min}_{k\in\{1\ldots C\}} S(\mathbf{X},\mathbf{Y}^*_k)
\label{eq:zero_shot_classification}
\end{equation}

\subsection{Hard negative mining}

In a metric learning problem, while the set of positive pairs $\mathbf{D_{+}}$ is fixed and given by the training set with one pair per positive image, the set of negative pairs $\mathbf{D_{-}}$ can be chosen more freely; indeed, as there are many more ways of being different than being equal, the number of negative and positive pairs may not be identical. Moreover, we will see that increasing the size of $\mathbf{D_{-}}$ compared to that of $\mathbf{D_{+}}$ by some factor $n$ leads to better overall results.  

In the following, we explore three different strategies to sample the distribution of negative pairs using several learning epochs: we first present a variant of the method of \cite{bucher:hal-01348827} and then describe two iterative greedy schemes.

\subsubsection{Random}
In \cite{bucher:hal-01348827}, negative pairs are obtained by associating a training image with an attribute vector chosen randomly among those of other seen classes, with one negative pair for each positive one. As a variant, we propose to generate randomly $n$ negative pairs (instead of one) for each positive pair, chosen as in \cite{bucher:hal-01348827}, \ie, by randomly sampling the set of attribute vectors from the other classes. We include in the objective function a penalization to compensate for the unbalance between positive and negative pairs (see section \ref{sec:criter}).

\subsubsection{Uncertainty}
This strategy is inspired by hard mining for object detection \cite{Fu2013cu,shrivastava2016training,li2013bootstrapping,canevet2014efficient} and consists in selecting the most informative negative pairs and iteratively updating the scoring function given by Eq.~(\ref{eq:score_attribute_embedding}). 
We denote by $S_{t}(\mathbf{X},\mathbf{Y})$ this score at time $t$. During training, each time step $t$ corresponds to a learning epoch. At the first epoch, $S_{1}(\mathbf{X},\mathbf{Y})$ is learned using the random negative pairs of \cite{bucher:hal-01348827}. At each time $t$ each pair of training image $\mathbf{X}_{i}$ and candidate annotation $\mathbf{Y}$ coming from different (but seen) classes is ranked according to the uncertainty score:  
\begin{equation}
u_t(\mathbf{Y}|{\mathbf{X}_{i}}) = \exp(-(S_{t}(\mathbf{X}_{i},\mathbf{Y})-S_{t}(\mathbf{X}_{i},\mathbf{Y}^*))) 
\end{equation}
where $\mathbf{Y}^*$ is the true vector of attributes of $\mathbf{X}_{i}$. The vector of attributes which are most similar to the actual one while coming from different classes are the most relevant for improving the model. We define a probability of generating the pair based on this similarity score and sample this distribution.

\subsubsection{Uncertainty/Correlation}
We propose to improve the previous approach by taking into account the intra-class correlation. The underlying principle governing the selection is that the most correlated vectors of attribute, in a given class, are the most useful ones to consider. The correlation can be measured by:
\begin{equation}
q(\mathbf{Y}) = \exp\left(\frac{-1}{|\mathcal{Y}_{k}|}\sum_{\mathbf{Y}'\in\mathcal{Y}_k} \left\Vert \mathbf{Y}-\mathbf{Y}'\right\Vert_2 \right)
\end{equation}
where $k$ is the true class index of $\mathbf{Y}$ and $\mathcal{Y}_k$ is the set of attribute vector representations.

A trade-off between uncertainty and correlation is obtained globally by using the following scoring function: 
\begin{equation}
p_t(\mathbf{Y} | \mathbf{X}_{i}) = u_t(\mathbf{Y} | \mathbf{X}_{i}) * q(\mathbf{Y})
\label{eq:uncertainty_correlation_score}
\end{equation}
where each image attribute vector $\mathbf{Y}$ at epoch $t$ has a score of $p_t$ to be associated with $\mathbf{X}_{i}$. The current set of negative pairs $\mathbf{D}_{-}$ at epoch $t$ is obtained by iteratively increasing the set with new data sampled according to Eq.~(\ref{eq:uncertainty_correlation_score}).

\subsection{Adaptation of the objective function \label{sec:criter}}
The original learning criterion~(\ref{eq:criterion_global_bucher}), such as defined in~\cite{bucher:hal-01348827}, assumes that negative and positive pairs are evenly distributed. This is not the case in the proposed approach: the criterion must be adapted to compensate for the imbalance between positive and negative pairs, by weighting the positive and negative pairs according to their frequencies:
\begin{equation}
\begin{split}
\mathcal{L}(\mathbf{W}_A,\mathbf{W}_X, \mathbf{b}_X, \tau) = \frac{1}{|\bf{D_{+}}|}\left(\sum_{i\in\mathcal{\bf{D_{+}}}} l_H(\mathbf{X_i},\mathbf{Y_i},Z_i,\tau) + \lambda l_A(\mathbf{X_i},\mathbf{Y_i},Z_i)\right) \\
+ \frac{1}{|\bf{D_{-}}|}\left(\sum_{j\in\mathcal{\bf{D_{-}}}} l_H(\mathbf{X_j},\mathbf{Y_j},Z_j,\tau)\right)
 + \mu R(\mathbf{W}_A,\mathbf{W}_X, \mathbf{b}_X)
\label{eq:criterion_global}
\end{split}
\end{equation}

\noindent This  criterion is updated at each new epoch when learning the model.

\section{Experiments\label{sec:exp}}

\subsubsection{Datasets} In this section we evaluate the proposed hard mining strategy on different challenging zero-shot learning tasks, by doing experiments on the 4 following public datasets: aPascal\&aYahoo (aP\&Y) \cite{farhadi2009describing}, Animals with Attributes (AwA) \cite{Lampert:2009ew}, CUB-200-2011 (CUB) \cite{Wah:2011vq} and SUN attribute (SUN) \cite{Patterson:2014cv} datasets. They have been designed to evaluate ZSC methods and contain a large number of categories (indoor and outdoor scenes, objects, person, animals, \etc) described using various semantic attributes (shape, material, color, part name \etc). To make comparisons with previous works possible, we used the same training/testing splits as  \cite{farhadi2009describing} (aP\&Y), \cite{Lampert:2009ew} (AwA), \cite{Akata:2015tv} CUB and  \cite{Jayaraman:2014wq} (SUN).

\subsubsection{Image features} For each dataset, we used the VGG-VeryDeep-19  \cite{simonyan2014very} CNN models, pre-trained on imageNet (without fine tuning) and extract the fully connected layer (\eg, FC7 4096-d) for representing the images. 

\subsubsection{Hyper-parameters}To estimate the three hyper-parameters ($\lambda$, the dimensionality of the metric space ($m$) and $\mu$) we apply a grid search validation procedure by randomly keeping 20\% of the training classes.
$\mathbf{W}_A$ and $\mathbf{W}_X$ are randomly initialized with normal distribution and optimized with  stochastic gradient descent.

\subsection{Zero-shot classification}

\begin{table}[tb]
\centering
\caption{Zero-shot classification accuracy (mean $\pm$ std) on 5 runs.  We report results with VGG-verydeep-19 \cite{simonyan2014very} features. unc./cor. = {\em Uncertainty/Correlation} method. The unc./cor. method can't be apply to the AwA dataset since all images of the same class have the same attributes, contrarily to the aP\&Y, CUB and SUN datasets.}
\label{acczstab}
\begin{tabular}{c|l|c|c|c|c|}
\hline
Feat.& Method & {aP\&Y} & {AwA} & {CUB} & {SUN}\\ \hline 
\hline
&Lampert \etal \cite{Lampert:2014fs} & 38.16 & 57.23 & - & 72.00 \\
&Romera-Paredes \etal \cite{romera2015embarrassingly}  & 24.22$\pm2.89$ & 75.32$\pm2.28$ & - & 82.10$\pm0.32$ \\
&Zhang \etal \cite{Zhang:2015vs}  & 46.23$\pm0.53$ & 76.33$\pm0.83$ & 30.41$\pm0.20$ & 82.50$\pm$1.32\\
&Zhang \etal \cite{zhang2016zero}  & 50.35$\pm2.97$ & { 80.46$ \pm0.53$} & 42.11$\pm0.55$ & 83.83$\pm$0.29\\
&Wang \etal \cite{wang2016zero} & - & 78.3 & \bf48.6$\bf\pm0.8$ & - \\
& Bucher \etal \cite{bucher:hal-01348827}   & 53.15$\pm0.88$ & { 77.32$\pm$1.03} & 43.29$\pm$0.38 & 84.41$\pm$0.71\\ \cline{2-6}
&Ours 'random' &{54.41$\pm1.47$} & {83.48$\pm0.99$} & {43.79$\pm0.68$} & {85.98$\pm1.14$} \\ 
\rot{\rlap{\small  VGG-VeryDeep}}\rot{\rlap{~~~~~~~\cite{simonyan2014very}}}
&Ours 'uncertainty' & 56.01$\pm0.58$ & {\bf86.55$\bf\pm1.07$} & {45.41$\pm0.10$} &  {\bf86.21$\bf\pm0.88$} \\ 
&Ours 'unc./cor.'&{\bf56.77$\bf\pm0.75$} & - & {45.87$\pm0.34$} & {86.10$\pm1.09$} \\ 
\hline
\end{tabular}
\end{table}

The experiments follow the standard ZSC protocol: during training, a set of images from known classes is available for learning the model parameters. At test time,  images from  unseen classes have to be assigned to one of the possible classes. Classes are described by a vector of attributes. Performance is measured by  mean accuracy and std over the classes.

Tables \ref{acczstab} and  \ref{randomnumberex} show the performance given by our hard-mining approach, which outperforms previous methods on 3 of the 4 datasets by more than 3\% on average (+9\% on AwA).  The smart selection of negative pairs plays a role on the decision boundaries  especially where classes have close attribute descriptions. 
We did not compare our results with \cite{Akata:2015tv} or \cite{xian2016latent} as they use different image features.
The 2 alternatives explored in the paper ({\em Uncertainty} vs {\em Uncertainty/Correlation}) give similar performance, but, as shown in the next section {\em Uncertainty/Correlation} is faster.

\subsection{Performance as a function of the ratio of positive/negative pair}

\begin{table}[tb]
\centering
\caption{Zero-shot classification accuracy (mean $\pm$ std) on aP\&Y dataset as a function of the ratio of positive/negative pairs.}
\label{randomnumberex}
\begin{tabular}{|l|l|l|l|l|}
\hline
Method / \#neg. pair & 1 & 10 & 50 & 100 \\ \hline\hline
Random  & 53.15$\pm0.88$  & 53.98$\pm0.79$   &   54.41$\pm1.47$  &  54.37$\pm1.05$  \\ \hline 
Uncertainty     & 55.47$\pm1.00$  &  55.84$\pm1.09$ & 55.48$\pm1.37$  &  56.01$\pm0.58$  \\   \hline
Uncertainty/Correlation                     & \bf56.08$\bf\pm0.41$  & \bf56.05$\bf\pm0.54$  & \bf56.69$\bf\pm1.78$   &  \bf 56.77$\bf\pm0.75$  \\   \hline
\end{tabular}
\end{table}

Table \ref{randomnumberex} give accuracy performances on aP\&Y dataset for the three methods in function of the number of negative examples for each positive pair. Bucher \etal \cite{bucher:hal-01348827} configuration corresponds to random method with one negative example per positive one. Our new negative pair selection method have a strong impact on the performance with a noticeable mean improvement of 3\%. Augmenting the ratio of negative pairs over positive ones has a positive influence on the accuracy. 

\subsection{Convergence}

\begin{figure}[tb]
\centering
\includegraphics[scale=0.4]{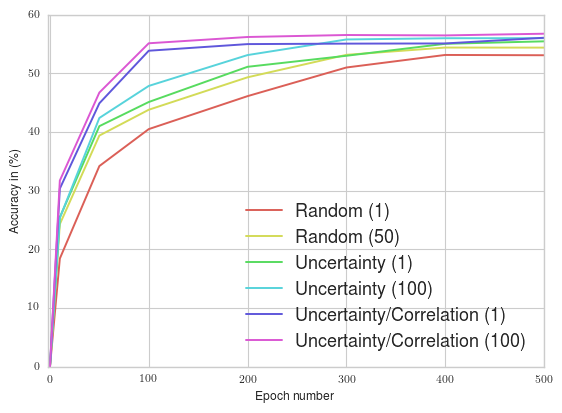}
\caption{Evolution of the performance as a function of the number of epochs,  on the aP\&Y dataset, with a neg/pos ratio of 1 and 100.}
\label{fig:accdim}
\end{figure}

We also made experiments to evaluate the impact of the hard mining selection on the convergence during training. Figure~\ref{fig:accdim} shows that {\em Uncertainty/Correlation} converges around 4 times faster than the {\em Uncertainty} and {\em Random} methods. This confirms the fact that more informative (negative) pairs are selected with this strategy. The negative/positive ratio has a (small) positive impact on the convergence.

\vspace{-.3cm}
\section{Conclusions}
This paper extended the original work of Bucher \etal \cite{bucher:hal-01348827} by proposing a novel hard negative mining approach used during training.
The proposed selection strategy gives close or above state-of-the-art performance on four standard benchmarks and has a positive impact on convergence.
\vspace{-.3cm}

\clearpage
\bibliographystyle{splncs}
\bibliography{0-bibfile.bib}

\end{document}